\documentstyle[ijcai89]{article}

\begin{document}

\title{
Natural Language Processing: Structure and Complexity}
\author{Wlodek Zadrozny \\
IBM Research,
T. J. Watson Research Center \\
Yorktown Heights, NY  10598 \\
\thanks{In: "Proc. SEKE'96,
8th Int. Conf. on Software Engineering and Knowledge
Engineering", Lake Tahoe, 1996, pages 595-602.}
{\tt wlodz@watson.ibm.com}}

\bibliographystyle{plain}

\maketitle

\begin{abstract}

We introduce a method for analyzing the complexity of natural language
processing tasks, and for predicting the difficulty new NLP tasks.

Our complexity measures are derived from the Kolmogorov complexity of
a class of automata --- {\it meaning automata}, whose
purpose is to extract relevant pieces of information from
sentences. Natural language semantics is defined only relative to the
set of questions an automaton can answer.

The paper shows examples of complexity estimates for various
NLP programs and tasks, and some recipes for complexity management.
It positions natural language processing as a subdomain of
software engineering, and lays down its formal foundation.

\end{abstract}

\section{Introduction}

This paper proposes a solution to the problem of measuring and
managing the complexity of natural language processing (NLP) systems.

Ideally, before building a NLP application, such as
a phone dialog system, a translation system, or a text skimmer,
we would like to know how difficult the task is. This
complexity measure
should be expressed in a number, which, in turn, could be
translated into an estimated program size (or other parameters),
and eventually
into dollars. If the task is too difficult, such a measure
would allow us to limit the task to subtasks of manageable
complexity; for instance, if a full natural language help
system for Unix is not feasible, perhaps
it is possible to
provide it for the twenty most common commands.

However, NLP domains and tasks are very seldom analyzed before
building a program, and typically very few numbers are provided
that would measure their complexity. For instance,
we can get information about the vocabulary size and the number
of grammar rules of a parser or a tagger, and subsequently
some numbers measuring its performance. But for NLP to become
an engineering discipline, we have to be able provide numbers
describing {\it in advance} the complexity of a task, e.g. of
parsing computer manuals.

For speech recognition some of those numbers have been computed,
e.g. the average perplexity of a corpus (roughly, the number of
possible continuations of the strings of words).
But we have not yet seen an analysis that would, for instance,
compute the number of different
sentences or dialogs expressing the command to
schedule a meeting with Bob at 5 in his office.

Such an analysis requires a
{\it model} of the phenomenon and {\it data}. For instance,
to measure the perplexity of a corpus,
we model the corpus as a set of sequences of words, and then the
counting produces the required numbers,
cf. \cite{Seneff92}.
Obviously, we could collect a corpus of data about meetings,
analyze it, and come up with the number. However, the data collection
process is costly, and thus cannot cover all
relevant domains. Furthermore, instead of counting possible dialogs,
it would clearly be easier to count language constructions that
appear in dialogs, and obtain the number of dialogs
by combinatorics.
This requires to model the language as a set of constructions, and thus
to abandon the simple counting model.

The natural question arises whether we can do better, and propose
a model that provides a measure
of the complexity of the task, but
dispenses with, or limits, the data collection effort.
We will propose such a model. It allows us to introduce several
measures of the complexity of NLP tasks; it
has a theoretical foundation in a modification of the theory of
Kolmogorov complexity; and it suggests ways of
managing the complexity of NLP programs by more careful
specification of their objectives.\\

\noindent
{\large\bf A digression on vision}

Instead of proceeding directly to the description of our model,
we will review some complexity estimates of problems in vision.
Our aim is to set up the stage for the discussion of complexity
issues in NL.

Moravec \cite{Moravec88,Moravec91} analyses the complexities
of different tasks, and compares them with the processing capacity
of computers, the human retina and the human brain.
Thus tracking a white line in perfect lighting conditions
requires about 1 MIP (million instructions per second);
smart bombs and simple road navigation take about 10 MIPS;
chess playing at grandmaster level --- 10,000 MIPS.
He estimates the computational capability of the human retina
at 1000 MIPS (which also
is the power of an IBM RS6000 workstation in 1995);
of the monkey visual system at 100,000 MIPS;
of the human visual system at 1,000,000 MIPS; and finally
of the human brain at 10 million MIPS.

These numbers are interesting, but it is more
important that this analysis is based on a specific model
In this model,
the processing powers are measured in {\it pixels/frame/second}.

To give an analogous complexity analysis of NLP,
we need a set of units in which we can measure the elements of
NLP tasks.

\section{Semantic complexity}

In this section we introduce the mathematical apparatus to measure
the complexity of NLP tasks.

In order to compare NLP systems, esp. if we want to compare systems
performing different tasks we
need a set of tools. One of the tools for
measuring complexity widely used in theoretical computer science is
{\it Kolmogorov complexity}. However, the concept of
K.-complexity must be modified to work for our task. This will be
done in this and the next section.

{\it Kolmogorov complexity} of a string $x$ is defined as the size of
the shortest string $y$ from which a certain universal Turing machine
produces $x$. Intuitively, $y$ measures the amount of information
necessary to describe $x$, i.e. the information content of $x$.
(cf. \cite{LiandVitanyi93} for details and a very good survey of
Kolmogorov complexity and related concepts).
For our purposes
any of the related definitions of complexity
will work. For example,
the Minimum Description Length of Rissanen
(\cite{LiandVitanyi93} and \cite{Rissanen82}), or
the size of a grammar (as in \cite{Savitch93}), or the
number of states of an automaton.

Obviously, K.-complexity does not by itself tell us anything about
natural language and semantics, so we will modify it, so that
we could define and measure the semantic complexity of natural language.
To do that, we assume that the meaning of a sentence is encoded
in the questions it can answer (examples below),
a view also held by Bloomfield
\cite{Bloomfield26}.

So, let $Q$ be a set of questions.
We will
define the {\it Q-complexity} of a set of sentences {\it S}
as the size of the simplest model scheme {\it M}$=M_S$,
such that, for any sentence {\it s}
in {\it S}, its semantics is given by {\it M(s)}, and {\it M(s)} correctly
answers all questions about {\it s} contained in $Q$.

The words "model scheme" can stand for either "Turing machine", or
"Prolog program", or "description", or a related notion. In this paper
we think of $M$ as a Turing machine that computes the
semantics of the sentences in $S$, and measure its size by the
product
{\it (Number of States)} $\times$ {\it (Number of Symbols)}. Thus defined,
$State \times Symbol$-complexity is a good
approximation of K-complexity (\cite{LiandVitanyi93}).

To define Q-complexity,
we first define the concept of {\it meaning automaton}
(M-automaton): Let, as above, $Q$ be a set of questions, and
$S$ set of sentences (e.g. accepted by a grammar).
Formally,
we treat each question as a (partial)
function from sentences to a set of answers $A$:
$$ q : S \rightarrow A $$
Intuitively, each question examines a sentence for a piece of
relevant information. Under this assumption the semantics of
a sentence (i.e. a formal string) is not given by its truth
conditions or denotation but by a set of answers:
$$  \| s \| = \{ < q, q( s ) > : q \in Q \} $$
Now, given a set of sentences $S$ and a set of questions $Q$, their
{\it meaning automaton} is a function
$$ M : S \times Q \rightarrow A $$
which satisfies the constraint
$$ M (s,q) =  q(s) $$
i.e. a function which gives a correct answer to every question.
We call it a meaning automaton because for any sentence $s$
$$  \| s \| = \{ < q,M(s,q) > : q \in Q \} $$
As before, the
{\it Q-complexity} of the set $S$ is the size of the smallest such $M$.\\

Note that the idea of a meaning automaton as a question
answer map allows us
to bypass all subtle semantics questions without
doing violence to them.

Note also that in practice we will deal not with
{\it the} simplest model scheme, but with the simplest we are
able to construct. Furthermore, to take care of the possible
non-computability of the function computing Q-complexity
of a set of sentences, we can put some restriction on
the Turing machine, e.g. requiring it to be finite state or
a stack automaton. Finally,
we will use the size of the Turing machine description
--- {\it T-rule--complexity} ---
as another, related, measure (see Sections 4 and 5).

\section{Some Q-complexity classes}

Meaning automata
(M-automata) provide a foundation for an
analysis of a few NLP problems and systems we
are going to present in this section. As a result of this analysis,
in the next sections, we should be able to present a method of
estimating the semantic complexities of NLP systems, and to suggest
ways of managing them.

We can measure the semantic complexity of a set of sentences
by the size of the smallest model that answers all relevant questions
about those sentences
(in practice, of the simplest one we are able to construct).
But what are the relevant questions?
Let us first look at the types of questions.
A simple classification of questions
given by \cite{QuirkSm73} (pp.191-2) is
based on the type of answer they expect:
(1) those that expect affirmation or rejection --- {\tt yes-no} questions;
(2) those that expect a reply supplying an item of information ---
 {\tt Wh} questions; and
(3) those that expect as the reply one of two or more options
presented in the question --- {\tt alternative} questions.

We will now examine a few simple Q-machines, and then discuss the
Q-complexities of a couple of programs.\\

\noindent
{\bf "what"-complexity}

Let $U$ be a finite set of tokens.
Consider the following semantic machine $M_U$: For any token $u$ in
$U$, if the input is "what is u" the output is a definition $def_u$
of $u$, and the next state is $acc$ (accept).
We assume that the output is just one token, $def_u$,
and the input also consists
only of one token, namely $u$, and the question is implicit. Then,
the size of $M_U$ is the measure of "what"-complexity of $U$.
$M_U$ can be described as a one state, two tape, Turing machine
consisting of the following set rules

$ (1, u, def_u, acc) $ for $u \in U $

\noindent
Both the T-rule-- \ and the $State \times Symbol$-- \ complexity
of $M_U$ are a simple function of the size of $U$.

This machine must be familiar to everyone; most keyword-based
help systems have this structure.\\

\noindent
{\bf "yes/no"-complexity}

In \cite{flairs96} we present a machine for answering
yes/no questions in two simple domains, each containing 24 sentences.
The machine simply compares the question with a stored on another tape
answer; it performs no analysis of the question except for extracting
a piece of data, and no reasoning. That machine had the
{\it State} $\times$ {\it Symbol} complexity of  $ 6 = 2 \times 3 $.

Obviously the size of the machine would grow if the number of
questions increases, or if some analysis is required.\\

\noindent
{\bf Q-Complexity of {\sc eliza}}

We can compute the semantic complexity of {\sc eliza}
\cite{Weizenbaum66} as Q-complexity.
Namely, we can identify $Q$
with the set of key-words for which {\sc eliza}
has rules for transforming input sentences (including a rule for
what to do if an input sentence contains no key-word).
{\sc eliza} had
not more than 2 key list structures for each of the about
50 keywords, and its control mechanism had 18 states.

If we assume that states and keywords corresponds to rules of a Turing
machine, we can estimate {\sc eliza}'s complexity at 118 rules.
Since the machine could be represented as a 4-column table, its T-rule
complexity would be roughly $4 \times 118 = 472$. (Its
$State \times Symbol$ complexity can be estimated in a similar way).

However, {\sc eliza} maintained the distinction between program
and data. Thus we can distinguish between the relative complexities of
{\sc eliza} database for the domain X, given {\sc eliza} program
(here about $4 \times 100$ for T-rules), and the complexity of
the program itself
(about $4 \times 18$). The moral is that all discussions of
complexity must assume some level of abstraction (see next section).
(Notice the parallels with the Lindenbaum algebras in logic).\\

\noindent
{\bf Complexity of extraction}

The importance of discussing semantic
complexity at some level of abstraction is even better seen
for more complex programs. We can turn our attention to
the area of information extraction
(e.g. \cite{muc93}). For instance,
Huffman, in a recent paper
\cite{huffman95}, uses about 60 extraction patterns generated
from 150 examples to create an NLP system attuned to
corporate management changes.
Thus the complexity of his system measured by the number of
patterns is 60.

This is the most relevant complexity measure for the system.
But at a different level of abstraction, its complexity would
be larger, since it would have to include the complexity(-ies) of the
grammar that is needed to break the text into noun and verb groups;
or even the complexity of determining the sentence boundaries.
And the binary representation of
the whole extraction machine would increase the
complexity by orders of magnitude. However, those lower level
complexities are irrelevant, because they do not directly reflect
the semantic of the task --- they have little to do with the set
$Q$.

{\bf Note}.
In addition to talking about Q-complexities at different abstraction
levels, we can also say that the number 60
reflects the complexity of the task at the level of 90\%
precision (as achieved by the program). This is natural
because of the possible translation between
Minimum  Description Length and K-complexity
(\cite{LiandVitanyi93}).

\section{Q-complexities of new tasks}

In our previous discussions we described the M-automata
at a certain natural level of abstraction. For instance, we have assumed
that the output of the automaton answering a "what" question consists
of only one token; or that the grammar rules used in the specification
of the Huffman's extraction machine can be taken for granted.
Since the intuitions about what constitutes such a natural level
of abstraction are often shared by researchers, M-automata can be used to
estimate the semantic complexity of existing programs.
Our next step
is to explain what might constitute such a "natural level
of abstraction", and to estimate its complexity as a function of $Q$, and
some properties of the task, e.g. the size of the domain.
This will allow us to estimate the complexity of NLP programs in
new domains.\\

\noindent
{\bf Natural levels of abstraction}

The Kolmogorov complexity of a task, i.e. the size of the smallest Turing
machine that achieves some required input-output behavior is not
computable. This leaves us with approximations, i.e. machines we
are able to construct. In building
such machines, we choose an architecture.
Even though many architectural
styles are possible, in our case, with
the experience of building NLP systems as knowledge
based programs, it is natural to assume that they can be decomposed
into the main control program and the knowledge base.
Thus the complexity of the whole system is a function of the
complexities of the two parts, the program and the knowledge base. \\

\subsection{The complexity of the knowledge base}

We view the knowledge base as a set of facts about the
categories of the domain.
In a limited domain, we find a
relatively small number of semantic categories. For example,
for the domain of calendars the number of concepts is less than 100;
for room scheduling it is about 20. Even for a relatively
complex office application, say, Lotus Notes, the number
of semantic categories is between 200 and 500 (the number depends
what counts as a category, and what is merely a feature).

Those background knowledge facts
describe properties that are relevant for the task.
E.g. if the task is scheduling meetings, and one of the concepts is
"range", specifying the
"beginning", "end", and "duration",
we also have to know
that the value of "beginning" is smaller than the value of "end".
We should also know that two different
meetings cannot occupy the same room in overlapping periods of time,
and the number of days in any month, and that meetings
are typically scheduled after the current date,
etc., etc.

The complexity of such a knowledge base is described by its size.
So, the natural question that arises is how many such facts we need
in the knowledge base. Is it billions? The next question is whether
they can be stated in a language that can be computationally
manipulated. After all, in principle,
any efficient representation of space could
require sensors and effectors.

Fortunately,
background knowledge is bounded, and for reasoning about small
domains propositional representations work fine;
there is evidence
(\cite{Graesser81}, \cite{Croth79},
\cite{ZadandJen91}) that the ratio
of the number of words to the number of facts necessary to understand
sentences with them is about 10:1. Second, experience suggests that this bound
works also for concepts in general, that is, it is not restricted
to linguistic knowledge.

This 10:1 bound makes the enterprise of building
natural language understanding systems for small domains
feasible. The
background knowledge about them can be organized, coded and
tested (cf. e.g. \cite{Zad94tlt}).\\

\subsection{Complexity of the control programs}

Dialog systems are a good example of
the NLP architecture in which the main task is decomposed into
the control program and the knowledge base.
The main algorithm of such programs
is quite simple: at each step the following
routine is called
\begin{verbatim}
Parse one sentence:
 1. Read sentence/string.
 2. Parse sentence using
    (a) grammar
    (b) background knowledge
    (c) contextual parameters
 3. Compute attributes important for application
 4. Update current context
 5. Reply
\end{verbatim}
The parser may consult with background
knowledge; the semantic interpreter
(point 3) uses the knowledge of the application and current context
(e.g. what the question was about) to interpret the string
(e.g. an answer that is a fragment with otherwise multiple
interpretations). In point 4, contextual variables are updated
(e.g. that the context does not contain a question, or that
the current action is moving a meeting); and in 2c,
contextual parameters might include the state of the application,
e.g. to choose one of many possible parses.

We refer to the choice of replies based on the computed attributes
of the input as {\em dialog management}. Clearly, a dialog manager
is the main control program of a dialog system. Thus the question
of the complexity of control program is the question about
the complexity of the dialog manager. Some of
the crucial things in the working of a dialog manager include:\\
(a) take an order, and figure it out ("set up an appointment");\\
(b) deal with parameters of the order ("in the afternoon");\\
(c) ask for parameters ("is it important?"); \\
(d) deal with a change in the parameters ("make it 6"); \\
(e) deal with a change in the order ("no, show my email"); \\
(f) answer simple questions ("do I have any meetings today?"); \\
(g) recover from speech recognition errors ("at what time,
two or eight?").\\

This list could be extended. Also, it is clear, that none of the
points (a)-(g) are among the properties of {\sc eliza}. The question now
is how complex a dialog machine we need for a useful behavior. How many
constraints should it satisfy? How many states are required?

Fortunately, the complexity of task oriented dialog machines is low.
Winograd and Flores (\cite{WinandFlo86}, p.64ff) argue that
the basic conversation for action machine needs only 9 states, and
8 basic actions
({\tt request, promise, counter, declare,
assert, reject, withdraw, renege}) that describe transitions between
those states.
This number is not universally agreed upon,
but other estimates are low too. For instance,
Bunt \cite{Bunt94}
in his classification lists 18 basic
dialog control functions and dialog acts.
One can of course argue about the adequacy of either model,
but the fact remains that, for simple tasks,
dialog complexity is limited by a small number of basic states.

\subsection
{\bf Complexities of NLP systems in small domains}

The above remarks about the simplicity of the control programs
apply to many other NLP systems, e.g. those for information extraction.
Our experience suggests that these estimates should work in small
domains. But what is {\it a small domain}?

A small domain should have
less than 300 concepts, because, given the 10:1 ratio of axioms to
concepts, and 3,000 axioms as the upper limit of the expert system
technology, we cannot expect to be able to handle more knowledge
about the domain. (Of course, we can imagine decomposing a larger
domain into weakly interacting small domains). Notice that
this postulate does not mean that we have to restrict ourselves e.g.
to less that 300 words in NLP applications --- such applications
usually deal with classes of words, i.e. concepts. For instance
in syntactic tagging,
the number of categories is usually less than 50 while the vocabulary
exceeds 100,000 tokens; {\sc eliza} has no limitation on
the vocabulary, but its list of concepts is limited to a few dozens.

Given the above constraints, in a small domain, there should be no need
to go beyond the 300 or so concepts. We can express this postulate
in terms of Q-complexity.
\footnote{
Note that domains are small only relative to a task. For example,
the domain of simple banking transactions (balances, payments,
transfers) is small if the task is to be able to order the execution
of a transaction; but it is large if we were to provide explanations
how the execution takes place, and why this way and not the other.}

\section{Estimating the complexity of NLP tasks}

In this section we discuss the semantic
complexities of background knowledge, grammars and control
programs for NLP tasks. The second problem seems to be to
large extent open, but, as we point out, there is a hope for
a solution based on Zipf's laws.

\subsection{An analysis of a dialog system}

We have started our work in dialog systems with {\sc mincal},
a meeting scheduler,
which we discuss briefly in the next section. Our latest system
is a simulated ATM machine, connected to a speech recognizer
and a text to speech system. The interesting fact about this system
is that we have
actually tried to estimate the complexity of the domain before
building the application.\\

\noindent
{\bf Semantic analysis of linguistic representation}

The system currently has three categories of actions: {\it executable},
{\it recognized}, and {\it unknown}.
The first category comprises of 9 actions,
each with 0-3 parameters:
{\tt transfer/3, withdraw/2, deposit/2, pay/1,
inquire/1, summary/1, ok/1} (possibly, a quitting action)
{\tt quit/0}, and {\tt help/0}. The category of non-executable
but recognized action consists of: {\tt cancel, past} (anything about
events), and {\tt time} (anything
referring to temporal data).
Everything else is {\tt unknown}.
The parameters of actions are among:
{\tt account, account\_to, account\_from},
and 5 types of {\tt bills}
({\tt mastercard, electricity, insurance, visa},
and {\tt phone}).

The semantic engine, which maps linguistic representations into
executable actions and their parameters consists of 110 prolog facts
and 20 rules. The facts are
used mostly to categorize linguistic constructions,
situations, and names of actions and parameters. For instance,
in this domain declarative sentences are all interpreted as potential
commands; "checking" is interpreted as a request for information
(in the absence of other context); and one action can be
represented by different verbs (synonyms).

The 20 rules describe how to get the actions names and their attributes
from linguistic representations.
The average size of a rule is about 12 predicates.

Clearly, the size of this knowledge base seems to confirm the
10:1 ratio of axioms to concepts we discussed in previous sections.\\

\noindent
{\bf Linguistic analysis}

Linguistic analysis is performed by a chart parser, which has
been used without any changes in all our applications. Thus,
its complexity can be factored out from our current discussion.

The grammar of the ATM dialog machine consists of about 600 lexical
entries and about 200 productions
In addition, there are 22
domain specific rules used to constrain the parser
(e.g. that nouns describing "utilities" do not modify each other,
in sentences such as  "pay phone visa and insurance").

Our first
rough estimate of the complexity of grammars for such small domains
was based on the 10:1 ratio. That is we assumed that we would have
10 synonyms for each lexical entry in a given class, and that
that the grammar would require adding about 40 new constructions
to the calendar grammar we used before, and that we would need
about 120 productions altogether. It turned out that indeed we
had to add about 40 new productions, but in addition, we had to
add about 40 domain dependent entries that behave like idioms.

Thus our initial estimate was not correct; and we hypothesize
in Section 5.3.
that Zipf's laws could be a better
way of estimating the size of a grammar.

\noindent
{\bf Dialog management}

There are  7 main states in the dialog management program
(the 5 states described in Section 4.2., plus "initialize"
and "end").
Each state has on average about 10 sub-states.
These numbers
agree with the estimates given in Section 4.2.

The knowledge base associated with the applications specifies
how to ask for a parameter depending on a situation. E.g.
whether to ask for a value, or suggest a value, or assume
a default. Thus its size is proportional to the number
of different situations, and its upper bound is the sum
of the powers
$2^{p_a}$ of parameters of actions $p_a$,
for all actions $a$ (not a large number if the
number of parameters is small).\\

\noindent
{\bf Executing actions}

An action to be executed must have the required parameters present.
For the actions in our domain the parameters are given either as
a list ("phone and insurance"), or as an attribute (e.g.
"all", coming from "pay all my bills").
The handling of the two cases is different, for instance, because
the reply to the user should be different in each case.
Altogether we have 21 cases of {\it execute\_action},
each described by an average of 8 clauses.

\subsection{A comparison of two dialog systems}

Using the tools introduced above we can analyze other NLP systems.
In this section, we compare two dialog systems:
{\sc mincal} (\cite{Coling94}), a calendar management program,
and {\sc boris} (\cite{Dyer83}, \cite{Lehnertetal83}), whose domain
of expertise was a divorce case:
\begin{verbatim}
-- Why did Sarah lose her divorce case?
-- She cheated on Paul.
\end{verbatim}

If we compare {\sc boris} (\cite{Dyer83}, \cite{Lehnertetal83})
with {\sc mincal} we notice some clear parallels.
First, they have an almost identical vocabulary size of about 350 words.
Secondly, they have a similar number of background knowledge facts.
Namely, {\sc boris} uses around 50 major knowledge structures such as
Scripts, TAUs, MOPs, Settings, Relationships etc.; on average,
the size of each such structure would not exceed 10 Prolog
clauses/axioms,
with
no more than 4 predicates with 2-3 variables each per clause,
if it were implemented in Prolog.
If we apply a similar metrics to {\sc mincal}, we get about 200 facts
expressing background knowledge about time, events and the calendar,
plus about 100 grammatical constructions, many of them dealing with
temporal expressions, others with agents, actions etc. Clearly then
the two systems are of about the same size. Finally, the main algorithms
do not differ much in their complexities (as measured by size and
what they do).\\

We now can explain the difference in the apparent
semantic complexities of their respective domains.
First, the vocabulary sizes and
the sizes of knowledge bases of
{\sc boris} and {\sc mincal}
are almost identical.
Thus, their "what"-complexities are roughly the same.

But we should note that for dialog systems it makes sense to talk
about {\it iterated Q-complexity}. That is, if the Q-complexity of one
round of dialog is X, then the complexity of 2 rounds might be
larger than X; e.g. if what-complexity of a help system is
200, measured by the number of items we can ask about at the very
beginning of a conversation, and
if we are permitted to ask what-questions
about any item in the answers,
the "twice iterated what-complexity" of the system
might be 3000.

Now, our theory can give an explanation of why the sentence
{\it The meeting is at 5} seems simpler than
{\it Sarah cheated on Paul}. Namely, for the last sentence we assume
not only the ability to derive and discuss
the immediate consequences of that fact
such as "broken obligation " or "is Paul aware of it?", but also
such related topics as "Sarah's emotional life" ,
"diseases", "antibiotics", and
"death of grandmother". In other words, the real complexity
of discussing a narrative is at least the complexity of
"iterated-what" combined with "iterated-why"
(and might as well include alternative questions).

By the arguments of the preceding section,
this would require
really extensive background knowledge, and
the Q-complexity (measured by the number of facts)
would range between $10^5$ and $10^7$;
assuming each round of dialog requires an iteration of
"why" and "what", with at least 5 rounds
of dialog, and an average of 10 facts per token in the database
of background knowledge.
Hence, the domain of the marital relations, defined by the ability
to discuss any relevant and related topic, is not a small domain.

In contrast, the Q-complexity of {\sc mincal} is less than
$1,000$ (measured again by the number of facts),
there are no restrictions on the number of exchanges;
but there is an implicit assumption that all dialogs are restricted
to the tasks of scheduling, canceling and moving meetings,
and there is no expectation of discussing the content of the
meetings with the machine (the meetings could have topics though).
Here, the set $Q$ consists of questions about
parameters of calendar events: {\it action\_name?},
{\it event\_time?, ...}.

Thus, the domain of
calendar action is manageable, the domain of divorce not.
Because the control programs for both domains are of roughly
the same complexity,
this difference has to attributed to the size of the required
database of background facts. The key to controlling the semantic
complexity lies in
limiting the interaction with the user to a clearly defined set
of (wh-) questions. And this is the topic of the next section.

\subsection{Estimating the grammar size}

Although our initial estimate of the size of the grammar was
wrong, the number
of constructions needed for the application was relatively small.

This is also true relative to a class of constructions. For instance,
for prepositional phrases constructions,
only a small percent of constructions with each preposition is
needed. For the task of scheduling a room
we need 5 out of 30 constructions with "for" mentioned in
Collins-Cobuild dictionary
(\cite{coco87});
for "from" the ratio is
3 out of 26;
for  "with"  it is 3/30; and for "at" 2/24.

This observation is not limited to prepositional phrases. The same
pattern holds for constructions with the verb "to be", "to have",
and many phrasal constructions. But notice that while the domain
selects constructions which makes sense there, the constructions
do not explicitly mention the domain. Thus they are reusable;
they encode general linguistic knowledge. \\

It seems that estimating the size of the grammar can be based
on {\it Zipf's laws} (\cite{Zipf49,Mandelbrot61,Shooman83}),
which specify distributions of linguistic items.
According to Zipf, if
we rank words in a given corpus
by their frequencies
(i.e. how often they appear in the corpus),
with the most frequent word receiving rank 1,
the next one rank 2, and so on, until the least frequent,
then it can be observed that the following equation holds:

{\it rank $\times$ frequency = constant}

\noindent
Zipf checked that this law (with different constants) applies to
words in many corpora and different languages, to the number of
meanings per word in a dictionary, to lengths of articles in different
editions of Encyclopedia Britannica, etc. That is, not only,
remarkably enough, the same functional relationship between ranks
and frequencies holds for words in different languages and different
corpora, but it also holds for other linguistic entities.

Zipf's laws appear in other situations and domains,
including software engineering (e.g. \cite{Shooman83}
\cite{NakanishiArano95}).
Furthermore, they apply both to syntactic and
semantic classifications. Thus
we can conjecture that they would apply
to English constructions in restricted domains.

Under these assumptions it should be possible to estimate
the size of the grammar for a NLP task given the number of
concepts in the task domain, required accuracy of processing,
and some domain-independent language parameters.
How exactly one would do it seems to be an open problem,
but Chapter 3 of
\cite{Shooman83} seems like a natural starting point.

\section{Complexity management}

From the point of view of building a natural language
dialog system, the commonsense
domain of divorce is too complex.
But the divorce domain can be managed, e.g. by restricting
the conversation to questions about
concrete parameters of divorce settlements.

How do we, in general, control the complexity of an NLP application?
Given the model we have introduced, there are four elements
in an NLP system: the conceptual domain,
the user, the NLP program, and the application
controlled by the program (e.g. a database). The management of
complexity starts with the definition of the domain. The number
of conceptual
entities determines the size of the database of background
knowledge, the size and complexity of the grammar, and the
complexity of the control program.

In part this is accomplished by
specifying the types of meanings that
can be dealt with by the system. The concept of Q-semantics and
meaning automata should help with that part of the analysis.
The result should be a bound on the combinatorics of interaction between
the user and the system.

The next element has to do with
the user understanding the capabilities of
the system. This includes, in its static part,
defining for the user
the range of acceptable topics and
the types of NL constructions, e.g. by forbidding
all "why" questions. The dynamic part
requires
the recognition and control of the borderline between what is
acceptable and what is not. For instance, a meeting scheduling system
can recognize a question about a past room assignment
only to reply that it
does not keep the database of old assignments.
The dynamic part must thus include the
recognition of the intentions of the user, and
for dialog systems, the
control of dialog (e.g. asking questions that require simple yes/no
replies).\\

Since $State \times Symbol$ complexity applies to machines at different
level of abstraction,
we can measure at least
four different types of complexities: \\
A. Relative to a task ("what"-, "why-", Q-);\\
B. Of the main engine;\\
C. Of the whole program; \\
D. Relative to desired performance.

Architectures are there to resolve the constraints of representation,
complexity, and performance; e.g. there are trade-offs between
the pipeline architecture for NLP and
the integrated syntax and semantics.
The choice of right representations
(abstract structures) allows us to keep the value of
A smaller than 10K even if the complexity
C of the whole program is greater than 100M states. Thus the management
of complexity requires thinking about the best architecture.
Furthermore, since it is possible to find examples where
small differences in D may correspond to large differences
in B and C (consider e.g. the difficulty of improving accuracy
of a search engine or a speech recognition program), the management
of the expectations of users and a clear definition of the task
is of great importance.

Finally, it should be noted that  although
the tools we have introduced to analyze problems
allow us to estimate the complexity of NLP programs in
new domains, we cannot make {\it exact} predictions of their
complexity. The complexity measures apply to mathematical models,
and they come close to the reality only if those models closely resemble
the reality.
This situation is an exact parallel of other engineering disciplines.

\section{Discussion}

We have introduced a methodology for predicting the difficulty of
coding new natural language processing
programs and for analyzing existing ones.
Our methods can also be used in
managing complexity of NLP applications by guiding possible
reformulations of the tasks.

The method is based on the idea of defining semantic complexity
with respect to a set of (abstract) questions, i.e. types of information
that the program is supposed to compute. Since those types of
information depend on a task, the complexity of the task can
be computed from the complexity of the question automaton.
Furthermore, this complexity is invariant with respect
to the representation language (because of the invariance of
the Kolmogorov complexity on which it is based).
\footnote{The existence
of such invariant complexity measures is
not obvious, for example, Simon in \cite{Simon81}, p.228, wrote
"How complex or simple a structure is depends critically upon the
way in which we describe it. Most of the complex structures found in
the world are enormously redundant, and we can use this redundancy to
simplify their description. But to use it, to achieve this
simplification, we must find the right representation".}

The paper defines (for NLP) a set of complexity measures, such as
the size of background knowledge, and different types of
wh-complexity (based on the types of questions). The complexities
can be measured in
{\it (Number of States)} $\times$ {\it (Number of Symbols)}, and
by the {\it Number of Axioms}, assuming a constant (equal to 10)
ratio of the number of axioms per concept.
The first measure is used for the complexity
of the control; the second for background knowledge.

This kind of analysis can be applied to other
related programs: both those documented in the literature,
e.g. \cite{Bilange92}, \cite{Wilenskyetal88},
\cite{Lehnertetal83}, \cite{Schank75}, and the small
application domains in which we tested our approach to dialog
management
(calendar, banking transactions, fast food, insurance and email).

We have also shown how
the complexities of dialog machines and background knowledge
can be computed
for knowledge-based NLP systems.
We have speculated that the complexity of the NLP grammar can
be computed based on Zipf's laws.

However, the paper is only the first step in the direction of
making quantitative assessments of the difficulty of building NLP
systems; and many of natural important questions remain open.
For instance,
the question about the size of the NL grammar per task; or
how to measure the
complexity of NLP tasks based on different architectures,
such as a cascade of finite state automata, or the traditional
morphology-syntax-semantics-pragmatics. \\
The final point in our discussion of the methods introduced in the paper
is their possible {\it relevance for software engineering}. Here we
would like to make the following points.
\begin{enumerate}
\item We have shown how
NLP can be thought of as a subdomain of software engineering.
\item The method of complexity analysis based on Q-automata
should apply to other programming tasks.
\item It is possible that the theory of Q-complexity
can provide the theoretical
justification for some software metrics.
\item We have shown that the complexity of NLP tasks can be estimated
at different levels of abstraction, and the same should hold for
any other programming task.
\end{enumerate}


\begin{thebibliography}{10}

\bibitem{Bilange92}
E.~Bilange and J-Y. Magadur.
\newblock A robust approach for handling oral dialogues.
\newblock {\em Proc. Coling'92}, pages 799--805, 1992.

\bibitem{Bloomfield26}
L.~Bloomfield.
\newblock A set of postulates for the science of language.
\newblock {\em Language}, vol.I, pages 799--805, 1926.
\newblock
Reprinted in: Hayden, D.E. et al. (Eds.).
{\it Classics in Linguistics}. Philosophical Library Inc., NY, 1967.

\bibitem{Bunt94}
H.~Bunt.
\newblock Context and dialogue control.
\newblock {\em Think}, 3(May):19--31, 1994.

\bibitem{Croth79}
E.J. Crothers.
\newblock {\em Paragraph Structure Inference}.
\newblock Ablex Publishing, Norwood, New Jersey, 1979.

\bibitem{Dyer83}
M.G. Dyer.
\newblock {\em In-Depth Understanding}.
\newblock MIT Press, Cambridge, MA, 1983.

\bibitem{Graesser81}
A.C. Graesser.
\newblock {\em Prose Comprehension Beyond the Word}.
\newblock Springer, New York, NY, 1981.

\bibitem{huffman95}
S.~B. Huffman.
\newblock Learning information extraction patterns from examples.
\newblock {\em Proc. IJCAI-95 workshop on New Approaches to Learning for Natural
  Language Processing}, 1995.

\bibitem{Lehnertetal83}
W.~Lehnert, M.G.Dyer, P.N.Johnson, C.J.Yang, and S.~Harley.
\newblock Boris -- an experiment in in-depth understanding of narratives.
\newblock {\em Artificial Intelligence}, 20(1):15--62, 1983.

\bibitem{LiandVitanyi93}
M.Li. and P.Vitanyi.
\newblock {\em An Introduction to Kolmogorov Complexity and Its Applications}.
\newblock Springer, New York, 1993.

\bibitem{Mandelbrot61}
B.~Mandelbrot.
\newblock On the theory of word frequencies and on related
Markovian models of   discourse.
\newblock In R.~Jakobson, editor, {\em Structure of Language and its
  Mathematical Aspects. Proc. Symp. in Applied Math. vol. XII}, pages 190--219,
  Providence, RI, 1961. American Mathematical Society.

\bibitem{Moravec88}
H.Moravec.
\newblock {\em Mind Children}.
\newblock Harvard University Press, Cambridge, MA, 1988.


\bibitem{Moravec91}
H.~Moravec.
\newblock Caution! robot vehicle!
\newblock In V.~Lifschitz, editor, {\em Artificial Intelligence and
  mathematical theory of of computation. Papers in honor of
J. McCarthy}, pages  331--344. 1991.

\bibitem{muc93}
{\em Proc. of the 5th Message Understanding Conference}.
\newblock Morgan Kaufman, 1993.

\bibitem{NakanishiArano95}
K.~Nakanishi and T.~Arano.
\newblock Class library maturity metric.
\newblock {\em NTT R\&D}, 44(1):9--14, 1995.

\bibitem{Rissanen82}
J.~Rissanen.
\newblock A universal prior for integers and estimation by minimum description
  length.
\newblock {\em Annals of Statistics}, 11:416--431, 1982.

\bibitem{QuirkSm73}
R.Quirk and S.Greenbaum.
\newblock {\em A Concise Grammar of Contemporary English}.
\newblock Harcourt Brace Jovanovich, Inc., New York, NY, 1973.

\bibitem{Savitch93}
W.~J. Savitch.
\newblock Why it might pay to assume that languages are infinite.
\newblock {\em Annals of Mathematics and Artificial Intelligence},
  8(1,2):17--26, 1993.

\bibitem{Schank75}
R.~C. Schank, editor.
\newblock {\em Conceptual Information Processing}.
\newblock Americal Elsevier, New York, NY, 1975.

\bibitem{Seneff92}
S.~Seneff.
\newblock Tina: A natural language system for spoken language application.
\newblock {\em Computational Linguistics}, 18(1):61--86, 1992.

\bibitem{Shooman83}
M.L. Shooman.
\newblock {\em Software Engineering}.
\newblock McGraw-Hill, New York, 1983.

\bibitem{Simon81}
H. Simon.
\newblock {\em The Sciences of the Artificial}.
\newblock MIT Press, Cambridge, MA, 1981.

\bibitem{coco87}
J. Sinclair \ (ed.).
\newblock {\em Collins-Cobuild English Language Dictionary}.
\newblock Collins ELT., London, 1987.

\bibitem{Weizenbaum66}
J.~Weizenbaum.
\newblock Eliza.
\newblock {\em Communications of the ACM}, 9(1):36--45, 1966.

\bibitem{Wilenskyetal88}
R.~Wilensky, D.N. Chin, M.~Luria, J.~Martin, J.~Mayfield, and D.~Wu.
\newblock The Berkeley Unix consultant project.
\newblock {\em Computational Linguistics}, 14(4):35--84, 1988.

\bibitem{WinandFlo86}
T. Winograd and F. Flores.
\newblock {\em Understanding Computers and Cognition}.
\newblock Ablex, Norwood, NJ, 1986.

\bibitem{Zad94tlt}
W. Zadrozny.
\newblock Reasoning with background knowledge -- a three-level theory.
\newblock {\em Computational Intelligence}, 10(2):150--184, 1994.

\bibitem{ZadandJen91}
W.Zadrozny and K.Jensen.
\newblock Semantics of paragraphs.
\newblock {\em Computational Linguistics}, 17(2):171--210, 1991.

\bibitem{flairs96}
W.~Zadrozny.
\newblock On the complexities of NLP systems.
\newblock In {\em Proc. FLAIRS'96,
The 9th Florida Artificial Intelligence Research Symposium}.
Key West, FL., 1996.

\bibitem{Coling94}
W. Zadrozny, M. Szummer, S. Jarecki, D.~E. Johnson, and L.
Morgenstern.
\newblock NL understanding with a grammar of constructions.
\newblock {\em Proc. Coling'94}, 1994.

\bibitem{Zipf49}
G.K. Zipf.
\newblock {\em Human Behavior and the Principle of Least Effort}.
\newblock Addison-Wesley Press, Cambridge,MA, 1949.

\end{thebibliography}
\end{document}